\documentclass[runningheads]{llncs}
\usepackage{graphicx}

\usepackage{tikz}
\usepackage{comment}
\usepackage{amsmath,amssymb} 
\usepackage{color}
\usepackage[accsupp]{axessibility}  

\usepackage[shortcuts]{extdash}
\usepackage{multirow}
\usepackage[pagebackref=true,breaklinks=true,colorlinks,bookmarks=false]{hyperref}
\usepackage[capitalize]{cleveref}
\usepackage[ruled,vlined]{algorithm2e}
\crefname{algocf}{Alg.}{Algs.}
\Crefname{algocf}{Algorithm}{Algorithms}
\newcommand{\NA}{---}
\begin{document}
\pagestyle{headings}
\mainmatter
\def\ECCVSubNumber{6727}  

\title{Implicit Neural Representations for Variable Length Human Motion Generation} 

\titlerunning{Implicit Motion Modeling}
\author{Pablo Cervantes\inst{1} \orcidID{0000-0002-5256-9317}
\and Yusuke Sekikawa\inst{2} \orcidID{0000-0003-1111-5949}
\and Ikuro Sato\inst{1,2} \orcidID{0000-0001-5234-3177} 
\and Koichi Shinoda\inst{1} \orcidID{0000-0003-1095-3203}} 
\authorrunning{P. Cervantes et al.}
\institute{Tokyo Institute of Technology \and
Denso IT Laboratory Inc.}
\maketitle

\begin{abstract}
We propose an action-conditional human motion generation method using variational implicit neural representations (INR).
The variational formalism enables action-conditional distributions of INRs, from which one can easily sample representations to generate novel human motion sequences.
Our method offers variable-length sequence generation by construction because a part of INR is optimized for a whole sequence of arbitrary length with temporal embeddings. 
In contrast, previous works reported difficulties with modeling variable-length sequences.
We confirm that our method with a Transformer decoder outperforms all relevant methods on HumanAct12, NTU-RGBD, and UESTC datasets in terms of realism and diversity of generated motions.
Surprisingly, even our method with an MLP decoder consistently outperforms the state-of-the-art Transformer-based auto-encoder.
In particular, we show that variable-length motions generated by our method are better than fixed-length motions generated by the state-of-the-art method in terms of realism and diversity.
Code at https://github.com/PACerv/ImplicitMotion.
\end{abstract}

\section{Introduction}
Generative models of human motion serve as a basis for human motion prediction \cite{barsoum2018hp,Battan_2021_WACV,honda2020rnn,chen2020comogcn,kanazawa2019learning,butepage2017deep}, human animation \cite{Starke2021,Starke2020}, and data augmentation for downstream recognition tasks \cite{meng2019sample,varol2021synthetic,doersch2019sim2real,STRAPS2020BMVC}.
There has been intensive research on generative models for realistic and diverse human motions \cite{yan2019convolutional,hou2020soul,li2021learn} and in particular methods that can generate motions while controlling some semantic factors such as emotion \cite{hou2020soul}, rhythm \cite{li2021learn} or action class \cite{petrovich21actor,chuan2020action2motion}.
For tasks such as rare action recognition, data-efficient action-conditional motion generation has great potential, since it may provide data augmentation even for rare actions.

For motion generation, the quality of generations is evaluated by their realism and diversity. 
Models need the ability to sample novel and rich representations to generate high-quality motions.
A suitable generative model yields distributions of representations in a latent space, where a simple distance measure corresponds to semantic similarity between motions so that interpolations provide novel and high-quality motions.
A common generative modeling approach is Variational Auto-Encoders (VAE) \cite{Kingma2014,Higgins2017betaVAELB,chuan2020action2motion,petrovich21actor}, which employ an encoder to infer a distribution from which representations of motions can be sampled and a decoder which reconstructs the data from the representation.
The reconstruction loss provides strong supervision, while the variational approach results in a representation space, which allows sampling of novel data with high realism and diversity.

Since human motions naturally vary in length depending on persons or action, it is important to consider variable lengths in motion generation.
For example, we would like the representations of quick (short) and slow (long) sitting motions to be different but closer to each other than the representation of a walking motion.
In RNN-based VAEs \cite{chuan2020action2motion}, representations are updated each time-step; thus, it is not obvious how to sample a particular action such as quick sitting.
Also, their recursive generation may accumulate error when generating long sequences.
In contrast, ACTOR, a Transformer VAE \cite{petrovich21actor}, should conceptually provide time-independent representations and generate variable-length motions without accumulating error.
Nevertheless, \cite{petrovich21actor} reports directly training with variable-length motions results in almost static motions, and accordingly ACTOR requires an additional fine-tuning scheme to enable variable-length motion generation.
It remains unclear what causes such issues with the Transformer architecture.

\begin{figure}[t]
\centering
\includegraphics[width=0.90\textwidth, trim=0cm 0cm 0cm 0cm, clip]{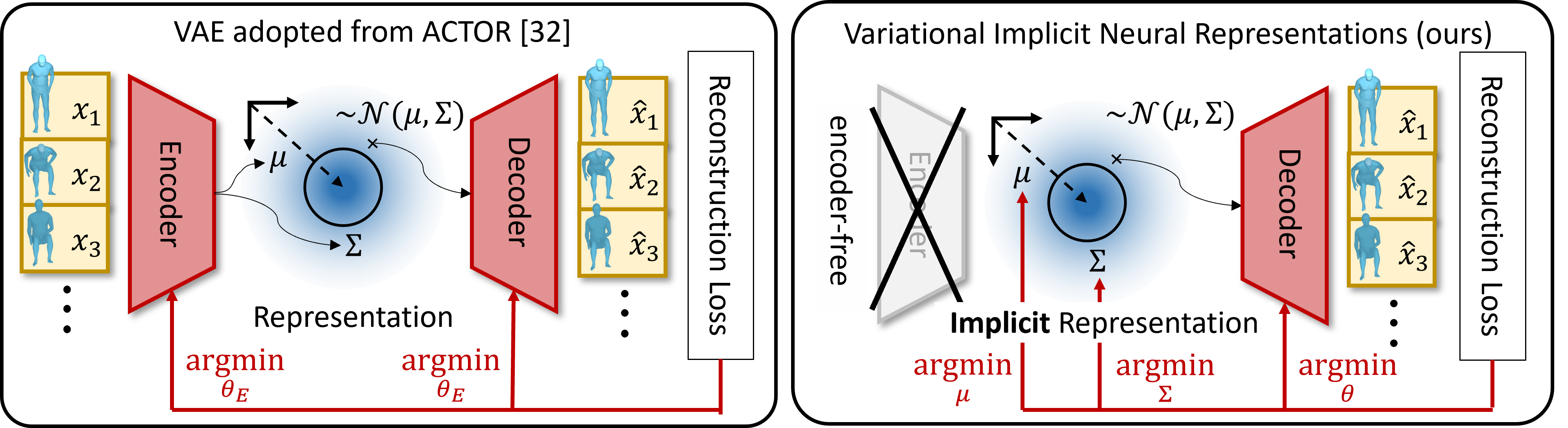}
\caption{Comparison between a Variational Auto-Encoder (VAE) baseline (top) and our variational implicit neural representation approach (bottom). In VAEs the encoder weights are optimized with respect to a full dataset and no guarantee of optimal representations for each individual sequence. In contrast, our sequence representations are directly optimized for each individual sequence and by construction offer variable-length sequence generation because a part of each INR is optimized for a whole sequence of arbitrary length. In this figure, we drop the temporal embeddings for simplicity.}
\label{fig:front_page_figure}
\end{figure}

A recently proposed generative modeling approach is Implicit Neural Representations (INR), which have been shown to be highly efficient in modeling complex data such as 3D scenes \cite{mildenhall2020nerf,OccupancyFlow,park2019deepsdf}.
INRs are representations that encode information without an explicit encoder, but through an optimization procedure as shown in \cref{fig:front_page_figure}.
INRs are usually constructed with respect to a decoder that takes a target coordinate and the representation of a target sample as input and returns the signal of the target sample at the target coordinate.
Such representations are optimized individually for each sample with respect to the reconstruction loss at all coordinates.
For a time-series, an INR is a time-independent, optimal representation that represents one whole sequence, regardless of the sequence length.
Since human motions are naturally variable-length, INRs are a very promising modeling approach.
However, to the best of our knowledge, there is no INR-based motion generation method that serves as a strong baseline.

To construct distributions from which one can sample a representation to generate a novel and high-quality motion, we propose variational INRs.
Compared to VAEs which infer variational distributions with an encoder, our variational INR framework models each sequence by a distribution with optimized parameters, e.g., mean and covariance, in the representation space.
We further decompose INRs into an action-wise part and sequence-wise part.
The action-wise part, whose distribution parameters are optimized for all sequences within the same action class, provides generalized components of an action.
The sequence-wise part, whose distribution parameters are optimized for an individual sequence, adds fine details of a specific sequence on top of the generalized components.

The average of the sequence distributions within an action class in the representation space serves as the action-specific generative model together with an appropriately trained decoder.
In our method, we further split the averaged distribution into several distributions depending on different intervals of sequence lengths.
This allows sampling of novel sequences with a target action and a target length.
To parameterize the action and sequence-length conditional distribution, we employ Gaussian Mixture Models (GMM).
Note that existing high-performing methods are unable to control sequence length.
This often results in poor motion generation with sequences ending before the action completes.

However, when fitting a GMM with a high degree of freedom to the representation space we risk simply reproducing training samples.
Previous evaluation metrics such as the Fr\'{e}chet Inception Distance (FID) and Diversity are not sensitive to this problematic model behavior, because they assign a high value to generated motions with a similar distribution as the training set.
In this regard, we propose a novel metric, the Mean Maximum Similarity (MMS), to measure such reproducing behavior.
By using this metric we confirm that our and previous studies successfully generate motions distinct from the training sequences.

We find that our proposed approach outperforms the current SOTA for action-conditional motion generation, ACTOR \cite{petrovich21actor}, in terms of realism and diversity.
By employing an identical decoder architecture as ACTOR \cite{petrovich21actor}, we conduct a fair comparison between our INR-approach and a VAE-approach and find that our INR approach improves motion generation.
Furthermore, since Transformer models can be difficult and expensive to train and we also explore the use of an MLP decoder and find that even such a simple, lightweight (6x fewer parameters) model can reach the SOTA performance.

Our key contributions are summarized as follows:
\begin{itemize}
    \item We propose a variational INR framework for motion generation, which gives time-independent, optimal representations for variable-length sequences distributed such that representations for novel motions can be easily sampled.
    \item To improve action-conditional motion generation, we propose INRs that are decomposed into action-wise and sequence-wise INRs. The action-wise INR generalizes to features across an action-class and helps generating realistic and novel motions for a target action class.
    \item We show in experiments that our method outperforms SOTA (ACTOR \cite{petrovich21actor}) on the HumanAct12, NTU13 and UESTC datasets in term of realism and diversity, and confirm that it generates high-quality variable-length sequences. For example on HumanAct12 we generate sequences with lengths between 8 - 470 time-steps and find that our motions generated with variable-length even outperform fixed-length motions generated by previous works (ACTOR\cite{petrovich21actor}, Action2Motion\cite{chuan2020action2motion}) in terms of realism and diversity.
\end{itemize}
\section{Related Works}
In the following we review the context of our work first regarding human motion modeling
and then regarding implicit neural representations.
\paragraph{Human Motion Modeling:}
The modeling of human motions is important for understanding and predicting human behavior.
Most modern approaches regard a human motion as a time-series of either skeleton poses or full 3D body shapes \cite{SMPL:2015,SMPL-X:2019} and previous works have proposed methods to estimate motions from videos, predicting future motions based on past motions and generating such motions conditional on signals such as emotion \cite{hou2020soul} and rhythm \cite{li2021learn}.
Our work is similar to \cite{petrovich21actor,chuan2020action2motion}, which generate motions conditional on the action class.

Previous works for motion generation are mostly based on Variational Auto-Encoders (VAE) \cite{Kingma2014,Higgins2017betaVAELB,chuan2020action2motion,petrovich21actor}, which employs an encoder to infer a variational distribution from which representations of motions can be sampled and a decoder which reconstructs data from a representation.
This encoder is optimized with respect to a reconstruction loss for a whole dataset, without a guarantee that the representation for each individual sequence is optimal.
The encoder may focus on the most common features in the dataset and become insensitive to rare features.
In contrast, Implicit Neural Representations (INR), which optimize the representation of each sample directly, can be sensitive even to unique features.

A similarity of all sequence modeling approaches is the use of model architectures such as RNNs or Transformers.
RNN are typically formulated as an auto-regressive model \cite{barsoum2018hp,chuan2020action2motion,hou2020soul}, which generates motions by recursively predicting the pose at time-step $t$ based on the prediction of the pose at time-step $t-1$.
This recursive nature of RNNs means that their sequence-representations are time-dependent and representations of variable-length sequences can not be easily compared.
Furthermore, the recursive generation procedure accumulates error and may result in poor performance when generating long sequences.

Petrovich et al. \cite{petrovich21actor} proposes ACTOR, a Transformer VAE, which yields a single fixed-length representation for a variable-length sequence through a Transformer encoder.
This representation is decoded by a Transformer decoder, which receives the representation and the temporal embedding of the target time-steps as input and generates the target sequence in one forward-pass.
Since such a Transformer VAE should conceptually handle variable-length sequences well, we choose this work as our main baseline.
However, \cite{petrovich21actor} reports that even for ACTOR a fine-tuning scheme is needed to enable good performance for variable-length sequences.

\paragraph{Implicit Neural Representations:}
INR as proposed in \cite{park2019deepsdf,OccupancyFlow,chen2018implicit_decoder} are encoder-free models which instead optimize their parameters to represent and fit a single sample. They have been popularized particularly in 3D modeling and have shown great performance on tasks such as inverse graphics \cite{mildenhall2020nerf,yariv2020multiview}, image synthesis \cite{Karras2021,Schwarz2020NEURIPS} or scene generation \cite{niemeyer2021giraffe,devries2021unconstrained}. While this work is, to the best of our knowledge, the first to explore implicit neural representations in the context of motion modeling, previous works have considered other time-series \cite{li2020neural,OccupancyFlow,sitzmann2019siren}.

Previous works for data synthesis using INR use a GAN-like approach \cite{Karras2021,anokhin2021image,Schwarz2020NEURIPS} for image synthesis. 
Such approaches don't optimize the INR, but sample representations from a predetermined distribution.
These representations are then used by a generator to generate images, which can fool a discriminator.
However, for our task the amount of training data required by GANs is problematic.

Another approach that takes inspiration from VAEs are variational INR \cite{park2019deepsdf,bond2020gradient}.
Most similar to ours is \cite{bond2020gradient}, however, this work doesn't optimize the INRs but rather approximates the optimal INR using an empirical Bayes.
Furthermore, they predict a variational distribution from the INR and then apply a regularizing loss to this intermediate representation instead of directly regularizing the INR.
Instead we directly optimize the mean and variance of the variational distribution as persistent parameters per sample.
The approach in \cite{park2019deepsdf} optimizes point-estimates of the representations of a sample and regularizes the distribution of these estimates, but doesn't sample stochastically from this distribution during training. Our work instead optimizes a distribution for each sample and samples stochastically from it during training.

With this work we would like to show that INR are not only powerful for high dimensional data such as dense 3D point clouds, but that their flexibility is also useful for other domains.
\section{Methodology}
\begin{figure}[t]
\centering
\includegraphics[width=\textwidth, trim=0cm 0cm 0cm 0cm, clip]{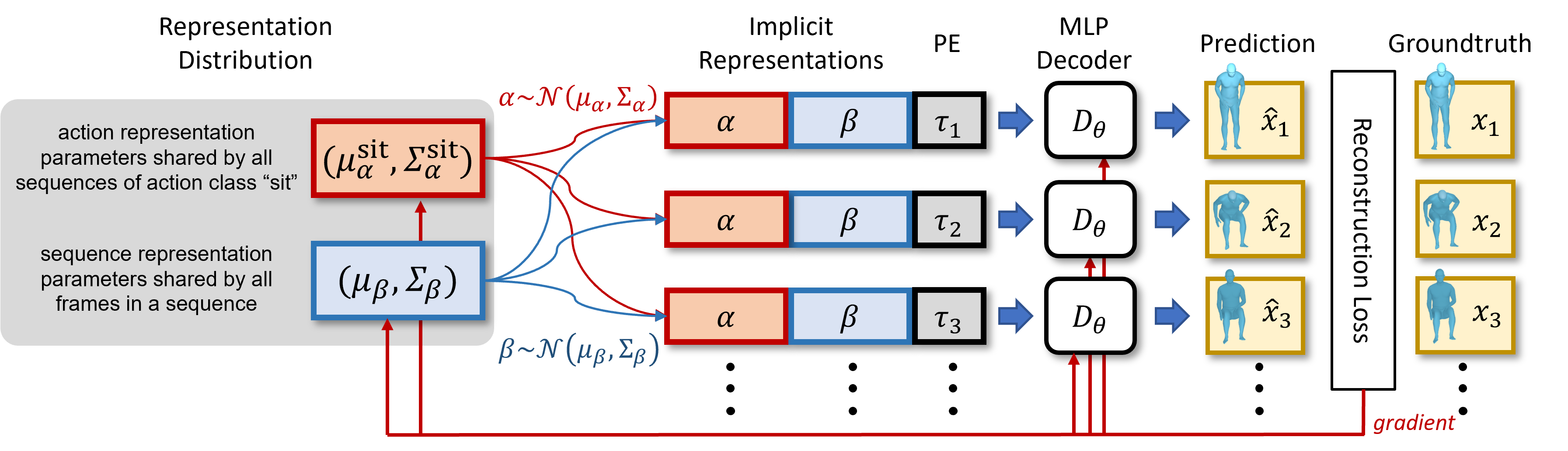}
\caption{Overview of Implicit Motion Modeling. Each representation is composed of two components, the action representation $\alpha$ and the sequence representation $\beta$. Instead of inferring these representations from an encoder, we directly optimize the parameters of a posterior normal distribution for both the action representation ($\mu_\alpha, \Sigma_\alpha$) shared by all sequences with the same action class and sequence representation ($\mu_\beta, \Sigma_\beta$). The representation, together with a temporal embedding (PE) $\tau_t$ of time $t$ is then input to an MLP, which predicts the pose at time $t$.}
\label{fig:main}
\end{figure}

In this section, we will first describe how to apply INRs to model human motions and decompose the INR into sequence-wise and action-wise representations (\ref{subsec:implicit_neural}).
Then we will introduce the proposed variational INRs (\ref{subsec:variational}), before we discuss how we fit a conditional Gaussian Mixture Model (GMM) to the representation space and how we sample novel sequence-wise representations from it (\ref{subsec:generative}).
Finally we will describe the Mean Maximum Similarity (MMS) as a measure to detect models that only reproduce training samples (\ref{subsec:mean_maximum_similarity}).

\subsection{Implicit Neural Representations for Motion Modeling}
\label{subsec:implicit_neural}
We consider a human motion as a sequence of poses represented by a low-dimensional skeleton.
Formally, we denote a skeleton pose of sequence $i$ at time $t$ as $x^i_t\in\mathbb{R}^{P \times B}$ where $P$ is the number of joints and $B$ is the dimensionality of the joint representation.
The $i$-th motion (a sequence of poses) is denoted as $\mathbf{x}^i = \{x_t^i\}_{t=1}^{T^i}$ with the sequence length $T^i$.

For each sequence $i$, we construct an Implicit Neural Representation (INR) $c_i$ and a decoder $D_\theta$ (shared among all sequences) that predicts a pose $\hat{x}_t^i$ of sequence $i$ from the INR $c_i$ and a temporal embedding $\tau_t$ of time $t$
\begin{equation}
    \hat{x}_t^i = D_\theta(c^i, \tau_t).
\end{equation}
Note that depending on the decoder architecture, the decoder may process all time-steps of a sequence independently (MLP) or multiple time-steps simultaneously (Transformer).
We obtain an INR $c_i$, shared by all time steps $(t\in\{1, 2, ..., T_i\})$, by minimizing the reconstruction loss $\mathcal{L}_\text{rec}^i$.
Thus, INRs can represent a sequence of any sequence-length $T^i$.
Also, for a given INR, the decoder can interpolate between time-steps (e.g. $t=0.5$) or extrapolate ($t>T_i$).

To generalize INRs to all features of the same action class, we decompose the INR and introduce an action representation shared across all samples of the same action class.
Formally, we divide each INR $c^i$ into a sequence-wise representation $\beta^i\in\mathbb{R}^S$ with in $i\in\mathcal{M}$ with a set of motions $\mathcal{M}$ and an action-wise representation $\alpha^{z}\in\mathbb{R}^A$ shared by all sequences with the same action label $z \in \mathcal{Z}$. Here $\mathcal{Z}$ is the set of action classes (e.g. $\alpha^z \in \{\alpha^\text{sit}, \alpha^\text{walk}, \alpha^\text{run}\dots\})$ and $S$ and $A$ denote the size of each representation respectively.

\subsection{Variational Implicit Neural Representations}
\label{subsec:variational}
Note that each INR $c^i$ is optimized to reconstruct a single sample with an over-parameterized decoder $D_\theta$.
This can make the distribution of INRs complex and result in a representation space where a simple distance measure  doesn't correspond to semantic similarity.
Accordingly, interpolations between representations in this space may not be meaningful.
To avoid such a complex representation space, we introduce a variational approach as regularization \cite{Kingma2014,Higgins2017betaVAELB}.
We formulate each INR as a normal distribution, whose mean $\mu^i$ and covariance matrix $\Sigma^i$ are optimized and from which we sample an instance with the re-parameterization trick during training.
This makes the representation space smoother so that close representations are semantically similar.
We summarize the sequence-wise and action-wise variational representations as
\begin{align}
\begin{split}
    c^i &\sim \mathcal{N}(\mu^i,\Sigma^i) \text{ with }\\
    \mu^i &= \text{concat}(\mu_\alpha^z, \mu_\beta^i),\\ 
    \Sigma^i &= 
    \begin{bmatrix}
        \Sigma_\alpha^z & 0^{A\times S}\\
        0^{S \times A}& \Sigma_\beta^i\\
    \end{bmatrix},
\end{split}
\end{align}
where $\text{concat}$ denotes the concatenation operation.

Furthermore, by assuming a standard normal distribution as the prior of each INR $c_i$, we further encourage a simple and compact representation space.
We then use the Kullback-Leibler (KL) Divergence $\mathcal{L}^i_{\text{KL}}$ as a regularizing loss
\begin{equation}
     \mathcal{L}_{\text{KL}}^i = \mathcal{D}_{\text{KL}}(\mathcal{N}(\mu^i,\Sigma^i)\|\mathcal{N}(0,I)).
\end{equation}
The sequence wise training objective of our method is thus
\begin{equation}
    \mathcal{L}^i = \mathcal{L}^i_\text{rec}+\lambda\mathcal{L}^i_{\text{KL}},
\end{equation} 
where $\mathcal{L}^i_{rec}$ is the reconstruction term
\begin{align}
    &\mathcal{L}_\text{rec}^i = -\mathbb{E}_{c^i\sim\mathcal{N}(\mu^i,\Sigma^i)}\sum_{t=1}^{T^i}\log p(x^i_t|c^i,\theta),\\
    \log& ~p(x^i_t|c^i,\theta) \propto \| x^i_t - D_\theta(\tau_t, \alpha^z, \beta^i) \|_2 + \text{const}., \nonumber
\end{align}
and $\mathcal{L}^i_{\text{KL}}$ is the regularizing KL divergence moderated by a weight $\lambda$.

We define the optimization problem for the model parameters as: 
\begin{eqnarray}
    \theta^\star = \underbrace{\underset{\theta}{\operatorname{argmin}}
    \sum_{z=1}^Z
    \underbrace{
    \underset{\mu_{\alpha}^z,\Sigma_{\alpha,}^z}{\operatorname{min}}
    \sum_{i\in\mathcal{M}^z}
    \underbrace{\underset{\mu_{\beta}^i,\Sigma_{\beta}^i}{\operatorname{min}}
    \mathcal{L}^i}_{\text{sequence-wise minimum}}}_{\text{action-wise minimum}}}_{\text{dataset-wise minimum}},
    \label{eq:model_train_objective}
\end{eqnarray}
where $\mathcal{M}^z$ denotes a set of sequence indices within action class $z$.
We optimize action-wise parameters $\mu_\alpha, \Sigma_\alpha$ for each action $z$:
\begin{equation}
   (\mu_\alpha^{z \star}, \Sigma_\alpha^{z \star}) = 
   \underset{\mu_{\alpha}^z, \Sigma_\alpha^z}{\text{argmin}}\sum_{i\in\mathcal{M}^z}
    \underset{\mu_{\beta}^i,\Sigma_{\beta}^i}{\operatorname{min}}
    \mathcal{L}^i.
    \label{eq:action_train_objective}
\end{equation}
Likewise, for the sequence-wise parameters we define the optimization problem for each sequence $i$ as:
\begin{equation}
    (\mu_\beta^{i \star},\Sigma_\beta^{i\star}) = \underset{\mu_{\beta}^i, \Sigma_{\beta}^i}{\text{argmin}} \mathcal{L}^i.
    \label{eq:sequence_train_objective}
\end{equation}
\subsection{Conditional GMM of Representation Space}
\label{subsec:generative}
To generate new sequences for a target action class, we need novel samples from the distribution of sequence-wise representations.
In the distribution of sequence-wise representations obtained during training, semantic factors such as sequence\-/lengths and action classes may be entangled.
Accordingly, the action-conditional distribution of sequence-wise representations may differ from the standard normal distribution.
To control sequence-length and action class for motion generation, we fit a conditional Gaussian Mixture Model (GMM) to the sequence-wise representations $\beta^i$ sampled 50 times from the variational distributions $\beta^i \sim \mathcal{N}(\mu_\beta^{i\star}, \Sigma_\beta^{i\star})$ for each training sequence.

We fit such a conditional GMM by first constructing subsets of sequence-wise representations that have the same action class $z$ and a sequence-length within the range $[T, T+\Delta T]$.
We choose the size of the sequence-length range $\Delta T$ to ensure a minimum number of samples in each subset and then fit an independent GMM to each subset of sequence-wise representations.
The details for how we select such a set of sequence-length ranges are provided in \cref{appendix:generativeModeling}.
Finally, we obtain the GMM of $p(\beta|z, [T, T+\Delta T])$.

To sample new sequence representations and generate corresponding novel motion sequences, we need to provide a target action class and sequence length.
We sample a new sequence representation $\beta^\text{new}$
\begin{equation}
     \beta^\text{new} \sim p(\beta|z, [T,T+\Delta T]),
\end{equation}
by sampling from the GMM corresponding to the target action class and sequence length.
With a new sequence representation we generate a new motion
\begin{equation}
    \mathbf{x}^\text{new} = \left\{D_{\theta^\star}(\alpha^{z\star}, \beta^\text{new}, \tau_t)\right\}_{t=t_0}^{T'}
\end{equation}
with the target action code $\alpha^{z\star}$ (obtained during the training stage) and the target sequence length $T' \in [T, T+\Delta T]$.

\subsection{Mean Maximum Similarity}
\label{subsec:mean_maximum_similarity}
By increasing the number of components of the GMM, it can better fit the training distribution, which improves the realism of generated motions.
However, we also risk fitting a GMM which only reproduces motions in the training set.
Previous metrics such as the Fr\'{e}chet Inception Distance (FID) or the Diversity compute the feature distribution of training and generated motions and compare these distributions.
Generated motions that have a similar distribution (FID) or variance (Diversity) as real motions are considered high-quality. Generated motions identical to the training set would be considered best by such metrics.

To detect models that just reproduce training samples, we introduce the Mean Maximum Similarity (MMS) as a complementary metric.
Similarly to previous metrics we extract the features from all training sample and generated motions.
Then for each generated motion, we find the training sample with the smallest feature distance (most similar) to it.
The mean distance over a large set of generated motions should be small for models that reproduce training samples and large for models that generate novel motions.
Formally we denote the features of a motion as $f$ and the sets of generated and training motion sequences $\mathcal{M}_\text{gen}$ and $\mathcal{M}_\text{train}$ respectively, and compute the MMS as

\begin{equation}
    \mathcal{D}_{MMS}(\mathcal{M}_\text{gen}, \mathcal{M}_\text{train}) = \frac{1}{|\mathcal{M}_\text{gen}|}\sum_{i \in \mathcal{M}_\text{gen}}\underset{j\in \mathcal{M}_\text{train}}{\min}(\|f_i - f_j\|_2).
    \label{eq:effective_distance}
\end{equation}

We estimate the MMS of model that only reproduces motions as baseline by computing $\mathcal{D}_{MMS}(\mathcal{M}_\text{train}, \mathcal{M}_\text{train})$ of the set of training motions $\mathcal{M}_\text{train}$ against itself. A large gap between $\mathcal{D}_{MMS}(\mathcal{M}_\text{gen}, \mathcal{M}_\text{train})$ and $\mathcal{D}_{MMS}(\mathcal{M}_\text{train}, \mathcal{M}_\text{train})$ indicates novel generated motions distinct from the training set.

\section{Experiments}
To verify the quality of motions generated by variational INR we perform experiments with a Transformer and an MLP decoder.
The Transformers is a powerful, but costly and difficult to train modeling tool, while the MLP is simple and comparatively light-weight.
The comparison should highlight the efficiency of the variational INR framework independent of decoder architecture.
In this section we will first explain the implementation details of our models (\ref{subsec:implementation}) and the datasets for our experiments (\ref{subsec:datasets}).
Then we will describe how we quantify the realism, diversity and novelty of generated motions (\ref{subsec:metrics}).
Finally we will discuss the quantitative (\ref{subsec:quantitative}) and qualitative (\ref{subsec:qualitative}) results.
\subsection{Implementation}
\label{subsec:implementation}
\paragraph{Skeleton representation:}
We represent the human body as a kinematic tree defined by joint rotations, bone-lengths and the root joint.
More specifically, we use the SMPL model \cite{SMPL:2015} with pose parameters consisting of 23 joint rotations, 1 global rotation and 1 root trajectory.
During training we only predict the pose parameters, which are independent of the body shape and can be used to animate any body at test time.
We represent rotations with a 6D rotation parameterization as proposed by \cite{zhou2019continuity} which means the full body pose has 147 dimensions ($24 \times 6 + 3$).
We use a reconstruction loss composed of a loss on the pose parameters (joint rotations and root joint locations) as well as the vertices of the SMLP model since \cite{petrovich21actor}'s findings suggest the best performance for this configuration.
On the NTU13 dataset, where at the time of writing the SMPL data was no longer available we represent the pose with a 6D rotation parameterization, but use a reconstruction loss on the joint locations (through forward-kinematics) as proposed by \cite{chuan2020action2motion} and find similarly high performance with our method.

\paragraph{Model Architecture:}
We implement our MLP decoder with ELU activations and 5 hidden layers (1000, 500, 500, 200, 100). The input are temporal embeddings with 256 dimensions and sequence-wise representations/action-wise representations, which are both 128 dimensions respectively, and the decoder outputs 147 dimensional pose parameters. This results in a network with 1,399,147 parameters. Due to the larger dataset size of UESTC we also implement a larger model (2000, 2000, 1000, 1000, 200, 100) with 8,265,147 parameters which is only used on UESTC.
We also implement a Transformer-decoder (same as ACTOR \cite{petrovich21actor}) with 8 layers, 4 attention heads, a dropout rate of 0.1 and a feedforward network of 1024 dimensions. With temporal embeddings with 256 dimensions and the same pose parameterization this results in a network with 8,465,299 parameters ($6\times$ more than the MLP model). More details are provided in \cref{A:training}.

Note that the Transformer-decoder is sensitive to the initialization of the implicit representations. If the variance parameters are initialized with a high variance the Transformer-decoder may fail to converge, while the MLP decoder is not sensitive to this phenomenon. We explore this more in \cref{appendix:initialization}.

The Transformer-based decoder has an identical structure to ACTOR \cite{petrovich21actor} and thus allows us a direct comparison between an auto-encoder and an implicit framework.
The MLP decoder is simpler to train than the Transformer decoder and doesn't rely on self-attention.
The comparison of these decoders allows us to determine if the choice of decoder architecture is critical for good performance.

\subsection{Datasets}
\label{subsec:datasets}
To evaluate the quality of action-conditional human motion generation, we used the UESTC, NTU-RGBD and HumanAct12 dataset curated by \cite{chuan2020action2motion}. \footnote{We considered the CMU Mocap dataset, but manual inspection found the label annotations for some actions such as ``Wash" and ``Step" to be extremely noisy.}
\paragraph{HumanAct12 \cite{chuan2020action2motion}:}
This dataset is based on PHSPD \cite{zou20203d} and consists of 1191 motion clips and 90099 frames in total.
Action labels for 12 actions are provided with at least 47 and at most 218 samples per label.
Sequence-lengths range from 8 to 470.
We follow the procedure by \cite{petrovich21actor} to align the poses to frontal view.
\paragraph{NTU13 \cite{liu2019ntu}:}
The NTU-RGBD dataset originally contains pose annotations from a MS Kinect sensor and label annotations for 120 actions.
\cite{chuan2020action2motion} re-estimated the data of a subset of 13 action, which we denote NTU13, with a state-of-the-art pose estimation method \cite{kocabas2020vibe} to reduce noise.
In this refined subset each action label has between 286 - 309 samples.
The refined poses have 18 body joints and the sequence lengths range from 20 - 201.
\footnote{Due to the release agreement of NTU RGBD, this subset can no longer be distributed. We report results to provide a complete comparison to previous studies.}
\paragraph{UESTC \cite{YanliUESTC}:}
This dataset with 40 action classes, 40 subjects and 25K samples is the largest dataset we perform experiments on and the only dataset with a train/test split.
We use the SMPL sequences provided by \cite{petrovich21actor} and apply the same pre-proprocessing, namely we rotate all sequences to frontal view.
Using the same cross-subject testing protocol we have a training split with between 225 - 345 samples per action class and sequence lengths between 24 and 2891 time steps (on average 300 time steps).

\subsection{Evaluation Metrics}
\label{subsec:metrics}
We use the same evaluation metrics as \cite{chuan2020action2motion,petrovich21actor} (Fr\'{e}chet Inception Distance (FID), action recognition accuracy, diversity and multimodality) to measure the realism and diversity of generated motions.
Also, we measure the proposed {\bf Mean Maximum Similarity} (\cref{subsec:mean_maximum_similarity}) to detect models that reproduce training samples.
We report a 95\% confidence interval computed of 20 evaluations.

The features for these evaluation metrics are extracted from motions of a predetermined length (60 time-steps) by an RNN-based action recognition model (weights provided by \cite{chuan2020action2motion}) for the NTU13 and HumanAct12 dataset and by an ST-GCN-based action recognition model (weights provided by \cite{petrovich21actor}) for the UESTC dataset. 
However, since the real training data is variable-length, we follow \cite{chuan2020action2motion}'s procedure for feature extraction during evaluation. This procedure adjust all sequences to a target length, by repeating the last pose of short sequences and by sampling random sub-sequence from longer sequences.  

Such stochastic feature extraction means the MMS may not be zero, even for sets of identical motions.
Thus we first compute a baseline MMS for identical real motions and then evaluate the MMS between real motions and generated motions.
If the MMS for generated motions is larger than that of real motions only, we conclude that the generated motions are distinct from the real motions.

We find that there is a difference in the evaluation procedure of previous works in the sampling frequency of different action classes for generation.
The approach by \cite{chuan2020action2motion} generates motions uniformly for all action classes.
On datasets with an action imbalance, this creates an inflated FID score.
We follow \cite{petrovich21actor}'s approach which generates motions according to the frequency of the action class in the training dataset, since this leads to more consistent results.

Our GMM samples novel representations conditional on the sequence-length.
For model evaluation we sample sequence-lengths according to their distribution in the training dataset.
We then sample corresponding representations and generate motions with the corresponding sequence-length.
We perform the same feature extraction as for the variable-length real motions.
More details can be found in the \cref{A:EvaluationMetrics}
\subsection{Quantitative Results}
\label{subsec:quantitative}
\setlength{\tabcolsep}{4pt}
\begin{table}[t]
\small
\begin{center}
\caption{Comparison on HumanAct12 and NTU13 (The best in bold, the second best underlined). \textit{Non-variational} uses action codes and \textit{no action code} uses the variational approach. ($\pm$ indicates 95\% confidence interval, $\rightarrow$ closer to real is better)}
\begin{tabular}{l c  c  c  c  c }
\hline\noalign{\smallskip}
\multirow{2}{*}{Method} & \multicolumn{4}{c}{HumanAct12}\\
& FID $\downarrow$ & Accuracy $\uparrow$& Diversity $\rightarrow$ & Multimod. $\rightarrow$\\ 
\hline\hline\noalign{\smallskip}
Real & $0.020^{\pm.010}$ & $0.997^{\pm.001}$ & $6.850^{\pm.050}$ & $2.450^{\pm.040}$\\
\hline\noalign{\smallskip}
Action2Motion \cite{chuan2020action2motion} &$0.338^{\pm.015}$ &$0.917^{\pm.003}$ &\underline{$6.879^{\pm.066}$} &\underline{$2.511^{\pm.023}$}\\
ACTOR \cite{petrovich21actor} &$0.12^{\pm.00}$ &$0.955^{\pm.008}$ &$\mathbf{6.84^{\pm.03}}$ &$2.53^{\pm.02}$\\
\hline\noalign{\smallskip}
INR (Transformer) & $\mathbf{0.088^{\pm.004}}$ & $\mathbf{0.973^{\pm.001}}$ & $6.881^{\pm.048}$ & $2.569^{\pm.040}$\\
INR (MLP) &\underline{$0.114^{\pm.001}$}&\underline{$0.970^{\pm.001}$}&$6.786^{\pm.057}$&$\mathbf{2.507^{\pm.034}}$\\
- (Non-variational) &$0.551^{\pm.005}$&$0.795^{\pm.002}$&$6.800^{\pm.046}$&$3.700^{\pm.032}$\\
- (No action code) &$0.146^{\pm.003}$& $0.955^{\pm.001}$& $6.797^{\pm.066}$&$2.769^{\pm.045}$\\
\hline\hline\noalign{\smallskip}
&\multicolumn{4}{c}{NTU13}\\
\hline\hline\noalign{\smallskip}
Real & $0.031^{\pm.004}$& $0.999^{\pm.001}$ & $7.108^{\pm.048}$& $2.194^{\pm.025}$\\
\hline\noalign{\smallskip}
Action2Motion \cite{chuan2020action2motion}&$0.351^{\pm.011}$&$0.949^{\pm.001}$&$\mathbf{7.116^{\pm.037}}$ &$\mathbf{2.186^{\pm.033}}$\\
ACTOR \cite{petrovich21actor}&\underline{$0.11^{\pm.00}$}&$0.971^{\pm.002}$&\underline{$7.08^{\pm.04}$} &$2.08^{\pm.01}$&\\
\hline\noalign{\smallskip}
INR (Transformer)&$\mathbf{0.097^{\pm..001}}$&$\mathbf{0.977^{\pm..001}}$&$7.060^{\pm.040}$&\underline{$2.108^{\pm.025}$}\\
INR (MLP)&\underline{$0.113^{\pm.001}$}&\underline{$0.976^{\pm.001}$}&$7.070^{\pm.052}$&$2.070^{\pm.043}$\\
- (Non-variational)&$0.646^{\pm.003}$&$0.849^{\pm.001}$&$6.905^{\pm.056}$&$3.244^{\pm.049}$\\
- (No action code)&$0.202^{\pm.002}$&$0.912^{\pm.001}$&$7.025^{\pm.050}$&$2.648^{\pm.043}$\\
\hline\noalign{\smallskip}
\end{tabular}
\label{tab:HumanAct12_NTU13}
\end{center}
\end{table}
\setlength{\tabcolsep}{1.4pt}
\begin{table*}[h]
\begin{center}
\def\arraystretch{1.3}
\small
\caption{Baseline comparison with UESTC. ($\pm$ indicates 95\% confidence interval, $\rightarrow$ closer to real is better}
\begin{tabular}{l c c c c c}
 \hline
Method & $\text{FID}_{\text{train}}\downarrow$ & $\text{FID}_{\text{test}}\downarrow$ & Accuracy $\uparrow$& Diversity $\rightarrow$ & Multimod. $\rightarrow$\\ 
\hline\hline
Real & $2.92^{\pm.26}$ & $2.79^{\pm.29}$ & $0.988^{\pm.001}$ & $33.34^{\pm.320}$ & $14.16^{\pm.06}$\\
\hline
ACTOR \cite{petrovich21actor} & $20.49^{\pm2.31}$ & $23.43^{\pm2.20} $ &$0.911^{\pm.003}$ & $\mathbf{31.96^{\pm.33}}$ & $\mathbf{14.52^{\pm.09}}$\\
\hline
Ours (MLP) & $\mathbf{9.55^{\pm.06}}$ & $\mathbf{15.00^{\pm.09}}$ &$\mathbf{0.941^{\pm.001}}$ & $31.59^{\pm.19}$ & $14.68^{\pm.07}$\\
\hline
\end{tabular}
\label{tab:UESTC}
\end{center}
\end{table*}

We compare our method to an RNN \cite{chuan2020action2motion} and a Transformer \cite{petrovich21actor} baseline and present some ablations for the proposed novel components of our model on HumanAct12 and NTU13 in \Cref{tab:HumanAct12_NTU13} and UESTC in \Cref{tab:UESTC}. 
Furthermore, we present a new state-of-the-art with the results for our Transformer-based and MLP-based models. 
We also investigate the contribution of variational INR by comparing them to a non-variational version and the contribution of the decomposed representations by comparing to a version with no action code.
Note that by construction our motion generation procedure can generate high-quality motions for arbitrarily specified sequence lengths (as in Table 1) within the variation of training sequence lengths, whereas previous works reported a performance drop for variable-length generation.

The results show that our proposed method improves over both Action2Motion \cite{chuan2020action2motion} and ACTOR\cite{petrovich21actor} especially on the FID and accuracy metric. We show that our optimized INR outperforms methods with representations produced by an optimized encoder. This is most apparent when comparing our implicit Transformer model and ACTOR, since both models use the same decoder architecture. 

Furthermore, we show high performance even with a simple MLP decoder. This shows that the self-attention mechanism is not necessary. We argue, that the common property of the Transformer-based models and our implicit MLP-based model, namely time-independent sequence-wise representations, are critical for motion generation performance. Such representations can avoid the error accumulation of current RNN-based models and represent variable-length sequences.

Comparing the non-variational and variational approach, we find that the realism and diversity is improved with the variational approach (see (Non-variational) in \Cref{tab:HumanAct12_NTU13}).
This finding suggests that the variational approach strongly regularizes the latent space and improves sampling of new motions.
However, both approaches for implicit representations are able to reach similar reconstruction performance for training samples and learn effective representations that allow high quality reconstruction.
\begin{table}
\def\arraystretch{1.3}
\footnotesize
\caption{Mean Maximum Similarity as a sanity check to detect overfitting.}
\begin{center}
    \begin{tabular}{l c c c}
        Method & HumanAct12 & NTU13 & UESTC\\
        \hline\hline
        Real &$0.329^{\pm.003}$&$0.209^{\pm.002}$&$4.925^{\pm.007}$\\
        Action2Motion\cite{chuan2020action2motion}&$0.945^{\pm.006}$&$0.667^{\pm.006}$&\NA\\
        ACTOR\cite{petrovich21actor} &$0.921^{\pm.001}$& $0.701^{\pm.001}$ &$8.645^{\pm 0.008}$\\
        INR (MLP) & $0.941^{\pm.005}$& $0.620^{\pm.001}$&$7.113^{\pm 0.006}$\\
        INR (Transformer) & $0.778^{\pm.003}$& $0.570^{\pm.002}$&\NA\\
        \hline
    \end{tabular}
    \label{tab:effectivedistance}
\end{center}
\end{table}

Comparing the approach with an action-wise and sequence-wise representation to an approach only with a sequence-wise representation, we find a clear advantage from using a decomposed action representation.
Even the approach with only sequence-wise representations performs comparable to the RNN-baseline (See (No action code) in \Cref{tab:HumanAct12_NTU13})
However, a decomposition representation is needed to outperform the Transformer baseline.

For further ablation studies we refer to \Cref{appendix:add_experiments}, where we investigate various modeling choices. Among others, we investigate the effect of the number of components in the GMM and show in \cref{appendix:gmm_components} high performance, on-par with Action2Motion\cite{chuan2020action2motion}, even when using GMMs with a single component.

Finally we perform a sanity check to see if any model is reproducing training samples by checking the proposed Mean Maximum Similarity between real and generated sequences defined by \cref{eq:effective_distance} and show the results in \Cref{tab:effectivedistance}.
We interpret the gap between this baseline of real motions and all other models as an indication that no model is just reproducing training samples.
Note that while the INR (Transformer) model has a lower MMS than other models, the gap to the baseline is significant. This suggests that it generates motions that are more similar to training motions than other models yet distinct from them.

\subsection{Qualitative Results}
\label{subsec:qualitative}
\begin{figure*}[t]
    \begin{center}
    \includegraphics[width=0.9\textwidth, trim=0cm 4.5cm 0cm 3cm, clip]{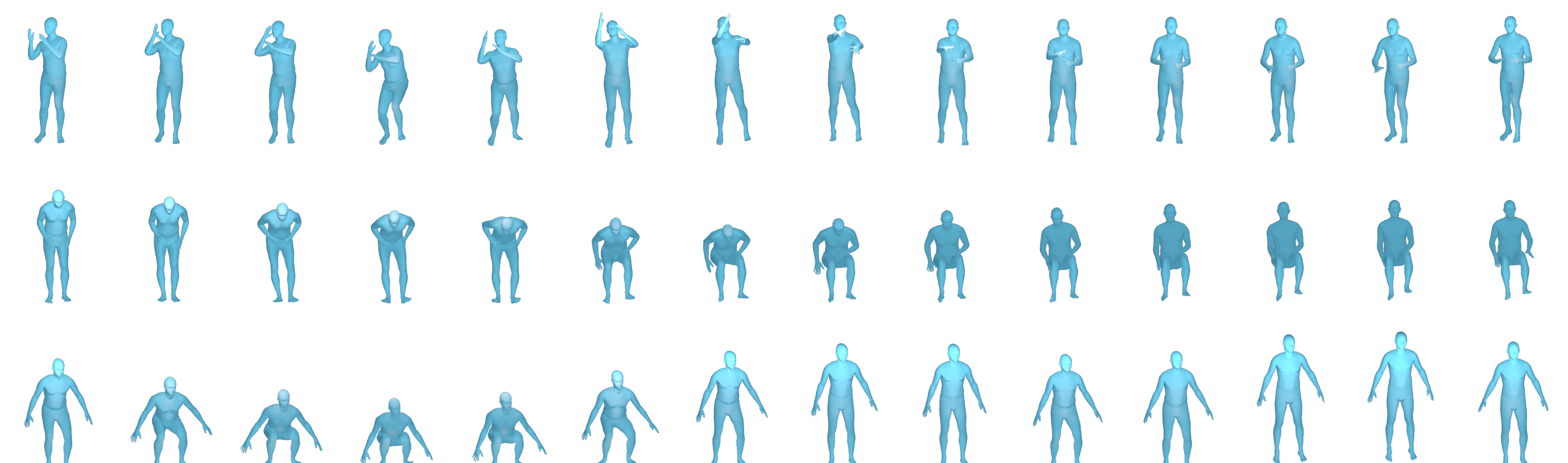}
    \caption{Motions generated by our MLP model trained on HumanAct predicted with 40 time-steps (each third frame shown) for the actions \textit{throw}, \textit{sit} and \textit{jump}.}
    \label{fig:Generated motions}
\end{center}
\end{figure*}

We manually inspect the quality of the generated motions and find that our methods consistently generates high-quality motions even over a long time range.
In particular we observe, that the RNN-based Action2Motion\cite{chuan2020action2motion} shows a slow-down effect when predicting long sequences, which neither our methods nor the Transformer-based ACTOR\cite{petrovich21actor} suffer from.

Also, we observe that we reliably generate complete actions, due to our model's ability to model the sequence-length.
In contrast, both Action2Motion as well as ACTOR tend to generate incomplete actions, particularly when generating short sequences. 
As shown in \cref{fig:Generated motions}, even for short sequences and actions with a clear start and end our generated actions are complete.
For more qualitative results, we refer to supplementary videos and \cref{A.AddQualitative}.

\section{Limitations and Future Work}
While our implicit sequence representations are parameter-efficient, the effort to train sequence-wise parameters scales linearly with the size of the training dataset.
Furthermore, we observe a sensitivity to the ratio of parameter updates between the sequence-wise parameters and the decoder parameters.
If the decoder parameters are updated significantly more often than sequence-wise parameters, our implicit models might perform poorly.
\section{Conclusion}
We present an MLP-based model for action-conditional human motion generation.
The proposed approach improves over previous RNN-based and Transformer-based baselines by employing variational implicit neural representations.
We argue that the likely reason for the success of our method is implicit neural representations, which are optimized representations that can represent full variable-length sequences and we supported this hypothesis experimentally by reaching state-of-the-art performance on commonly used metrics for motion generation.
While the results of our work may improve technologies for animation and action recognition, there is the potential for malicious use such a deep fakes as well.
For more detailed discussions about such potential negative societal impacts and personal data of human subjects we refer to \cref{appendix:NegativeImpact,appendix:Privacy}.

\textbf{Acknowledgements:}
This work is an outcome of a research project, Development of Quality Foundation for Machine-Learning Applications, supported by DENSO IT LAB Recognition and Learning Algorithm Collaborative Research Chair (Tokyo Tech.). It was also supported by JST CREST JPMJCR1687.
\bibliographystyle{splncs04}
\bibliography{egbib}

\begin{thebibliography}{10}
\providecommand{\url}[1]{\texttt{#1}}
\providecommand{\urlprefix}{URL }
\providecommand{\doi}[1]{https://doi.org/#1}

\bibitem{anokhin2021image}
Anokhin, I., Demochkin, K., Khakhulin, T., Sterkin, G., Lempitsky, V.,
  Korzhenkov, D.: Image generators with conditionally-independent pixel
  synthesis. In: Proceedings of the IEEE/CVF Conference on Computer Vision and
  Pattern Recognition (CVPR). pp. 14278--14287 (2021)

\bibitem{barsoum2018hp}
Barsoum, E., Kender, J., Liu, Z.: {HP-GAN}: Probabilistic {3D} human motion
  prediction via {GAN}. In: Proceedings of the IEEE conference on computer
  vision and pattern recognition workshops (CVPRW). pp. 1418--1427 (2018)

\bibitem{Battan_2021_WACV}
Battan, N., Agrawal, Y., Rao, S.S., Goel, A., Sharma, A.: {GlocalNet}:
  Class-aware long-term human motion synthesis. In: Proceedings of the IEEE/CVF
  Winter Conference on Applications of Computer Vision (WACV). pp. 879--888
  (2021)

\bibitem{bond2020gradient}
Bond{-}Taylor, S., Willcocks, C.G.: Gradient origin networks. In: 9th
  International Conference on Learning Representations, {ICLR} 2021, Virtual
  Event, Austria, May 3-7, 2021. OpenReview.net (2021)

\bibitem{butepage2017deep}
Butepage, J., Black, M.J., Kragic, D., Kjellstrom, H.: Deep representation
  learning for human motion prediction and classification. In: Proceedings of
  the IEEE/CVF Conference on Computer Vision and Pattern Recognition (CVPR).
  pp. 6158--6166 (2017)

\bibitem{chen2020comogcn}
Chen, Y., Liu, C., Shi, B.E., Liu, M.: {CoMoGCN}: Coherent motion aware
  trajectory prediction with graph representation. In: British Machine Vision
  Conference (BMVC) (2020)

\bibitem{chen2018implicit_decoder}
Chen, Z., Zhang, H.: Learning implicit fields for generative shape modeling.
  Proceedings of the IEEE/CVF Conference on Computer Vision and Pattern
  Recognition (CVPR)  (2019)

\bibitem{devries2021unconstrained}
DeVries, T., Bautista, M.A., Srivastava, N., Taylor, G.W., Susskind, J.M.:
  Unconstrained scene generation with locally conditioned radiance fields. In:
  Proceedings of the IEEE/CVF International Conference on Computer Vision
  (ICCV). pp. 14304--14313 (2021)

\bibitem{doersch2019sim2real}
Doersch, C., Zisserman, A.: Sim2real transfer learning for 3d human pose
  estimation: motion to the rescue. Advances in Neural Information Processing
  Systems (NeurIPS)  \textbf{32} (2019)

\bibitem{chuan2020action2motion}
Guo, C., Zuo, X., Wang, S., Zou, S., Sun, Q., Deng, A., Gong, M., Cheng, L.:
  {Action2Motion}: Conditioned generation of {3D} human motions. In:
  Proceedings of the 28th ACM International Conference on Multimedia (MM '20)
  (2020)

\bibitem{Higgins2017betaVAELB}
Higgins, I., Matthey, L., Pal, A., Burgess, C., Glorot, X., Botvinick, M.,
  Mohamed, S., Lerchner, A.: $\beta$-vae: Learning basic visual concepts with a
  constrained variational framework. In: 5th International Conference on
  Learning Representations, {ICLR} 2017, Toulon, France, April 24-26, 2017,
  Conference Track Proceedings. OpenReview.net (2017)

\bibitem{honda2020rnn}
Honda, Y., Kawakami, R., Naemura, T.: Rnn-based motion prediction in
  competitive fencing considering interaction between players. In: British
  Machine Vision Conference (BMVC) (2020)

\bibitem{hou2020soul}
Hou, Y., Yao, H., Sun, X., Li, H.: Soul dancer: Emotion-based human action
  generation. ACM Transactions on Multimedia Computing, Communications, and
  Applications (TOMM)  \textbf{15}(3s),  1--19 (2020)

\bibitem{YanliUESTC}
Ji, Y., Xu, F., Yang, Y., Shen, F., Shen, H.T., Zheng, W.S.: A large-scale
  {RGB-D} database for arbitrary-view human action recognition. In: Proceedings
  of the 26th ACM International Conference on Multimedia (MM '18) (2018)

\bibitem{kanazawa2019learning}
Kanazawa, A., Zhang, J.Y., Felsen, P., Malik, J.: Learning {3D} human dynamics
  from video. In: Proceedings of the IEEE/CVF Conference on Computer Vision and
  Pattern Recognition (CVPR) (2019)

\bibitem{Karras2021}
Karras, T., Aittala, M., Laine, S., H\"{a}rk\"{o}nen, E., Hellsten, J.,
  Lehtinen, J., Aila, T.: Advances in Neural Information Processing Systems
  (NeurIPS)  \textbf{34} (2021)

\bibitem{Kingma2014}
Kingma, D.P., Welling, M.: Auto-encoding variational bayes. In: 2nd
  International Conference on Learning Representations, {ICLR} 2014, Banff, AB,
  Canada, April 14-16, 2014, Conference Track Proceedings (2014)

\bibitem{kocabas2020vibe}
Kocabas, M., Athanasiou, N., Black, M.J.: Vibe: Video inference for human body
  pose and shape estimation. In: Proceedings of the IEEE/CVF Conference on
  Computer Vision and Pattern Recognition (CVPR). pp. 5253--5263 (2020)

\bibitem{li2021learn}
Li, R., Yang, S., Ross, D.A., Kanazawa, A.: Ai choreographer: Music conditioned
  3d dance generation with {AIST++}. In: Proc. of the IEEE International Conf.
  on Computer Vision (ICCV) (2021)

\bibitem{li2020neural}
Li, Z., Niklaus, S., Snavely, N., Wang, O.: Neural scene flow fields for
  space-time view synthesis of dynamic scenes. In: Proceedings of the IEEE/CVF
  Conference on Computer Vision and Pattern Recognition (CVPR) (2021)

\bibitem{liu2019ntu}
Liu, J., Shahroudy, A., Perez, M., Wang, G., Duan, L.Y., Kot, A.C.: {NTU RGB+D}
  120: A large-scale benchmark for 3d human activity understanding. IEEE
  transactions on pattern analysis and machine intelligence  \textbf{42}(10),
  2684--2701 (2019)

\bibitem{SMPL:2015}
Loper, M., Mahmood, N., Romero, J., Pons-Moll, G., Black, M.J.: {SMPL}: A
  skinned multi-person linear model. ACM Trans. Graphics (Proc. SIGGRAPH Asia)
  \textbf{34}(6),  248:1--248:16 (Oct 2015)

\bibitem{meng2019sample}
Meng, F., Liu, H., Liang, Y., Tu, J., Liu, M.: Sample fusion network: An
  end-to-end data augmentation network for skeleton-based human action
  recognition. IEEE Transactions on Image Processing  \textbf{28}(11),
  5281--5295 (2019)

\bibitem{mildenhall2020nerf}
Mildenhall, B., Srinivasan, P.P., Tancik, M., Barron, J.T., Ramamoorthi, R.,
  Ng, R.: {NeRF}: Representing scenes as neural radiance fields for view
  synthesis. In: European conference on computer vision (ECCV). pp. 405--421.
  Springer (2020)

\bibitem{niemeyer2021giraffe}
Niemeyer, M., Geiger, A.: {GIRAFFE}\: Representing scenes as compositional
  generative neural feature fields. In: Proceedings of the IEEE/CVF Conference
  on Computer Vision and Pattern Recognition (CVPR) (2021)

\bibitem{OccupancyFlow}
Niemeyer, M., Mescheder, L., Oechsle, M., Geiger, A.: Occupancy flow: {4D}
  reconstruction by learning particle dynamics. In: Proceedings of the IEEE/CVF
  International Conference on Computer Vision (ICCV) (2019)

\bibitem{park2019deepsdf}
Park, J.J., Florence, P., Straub, J., Newcombe, R., Lovegrove, S.: {DeepSDF}\:
  Learning continuous signed distance functions for shape representation. In:
  Proceedings of the IEEE/CVF Conference on Computer Vision and Pattern
  Recognition (CVPR). pp. 165--174 (2019)

\bibitem{paszke2019pytorch}
Paszke, A., Gross, S., Massa, F., Lerer, A., Bradbury, J., Chanan, G., Killeen,
  T., Lin, Z., Gimelshein, N., Antiga, L., et~al.: Pytorch: An imperative
  style, high-performance deep learning library. Advances in Neural Information
  Processing Systems (NeurIPS)  \textbf{32} (2019)

\bibitem{SMPL-X:2019}
Pavlakos, G., Choutas, V., Ghorbani, N., Bolkart, T., Osman, A.A.A., Tzionas,
  D., Black, M.J.: Expressive body capture: {3D} hands, face, and body from a
  single image. In: Proceedings of the IEEE/CVF Conference on Computer Vision
  and Pattern Recognition (CVPR) (2019)

\bibitem{scikit-learn}
Pedregosa, F., Varoquaux, G., Gramfort, A., Michel, V., Thirion, B., Grisel,
  O., Blondel, M., Prettenhofer, P., Weiss, R., Dubourg, V., Vanderplas, J.,
  Passos, A., Cournapeau, D., Brucher, M., Perrot, M., Duchesnay, E.:
  Scikit-learn: Machine learning in {P}ython. Journal of Machine Learning
  Research  \textbf{12},  2825--2830 (2011)

\bibitem{petrovich21actor}
Petrovich, M., Black, M.J., Varol, G.: Action-conditioned 3{D} human motion
  synthesis with {T}ransformer {VAE}. In: International Conference on Computer
  Vision (ICCV) (2021)

\bibitem{ravi2020pytorch3d}
Ravi, N., Reizenstein, J., Novotny, D., Gordon, T., Lo, W.Y., Johnson, J.,
  Gkioxari, G.: Accelerating {3D} deep learning with pytorch3d.
  arXiv:2007.08501  (2020)

\bibitem{Schwarz2020NEURIPS}
Schwarz, K., Liao, Y., Niemeyer, M., Geiger, A.: {GRAF}\: Generative radiance
  fields for 3d-aware image synthesis. In: Advances in Neural Information
  Processing Systems (NeurIPS) (2020)

\bibitem{STRAPS2020BMVC}
Sengupta, A., Budvytis, I., Cipolla, R.: Synthetic training for accurate 3d
  human pose and shape estimation in the wild. In: British Machine Vision
  Conference (BMVC) (September 2020)

\bibitem{sitzmann2019siren}
Sitzmann, V., Martel, J.N., Bergman, A.W., Lindell, D.B., Wetzstein, G.:
  Implicit neural representations with periodic activation functions. In:
  Advances in Neural Information Processing Systems (NeurIPS) (2020)

\bibitem{Starke2020}
Starke, S., Zhao, Y., Komura, T., Zaman, K.: Local motion phases for learning
  multi-contact character movements. ACM Trans. Graph.  \textbf{39}(4) (Jul
  2020)

\bibitem{Starke2021}
Starke, S., Zhao, Y., Zinno, F., Komura, T.: Neural animation layering for
  synthesizing martial arts movements. ACM Trans. Graph.  \textbf{40}(4) (Jul
  2021)

\bibitem{varol2021synthetic}
Varol, G., Laptev, I., Schmid, C., Zisserman, A.: Synthetic humans for action
  recognition from unseen viewpoints. International Journal of Computer Vision
  \textbf{129}(7),  2264--2287 (2021)

\bibitem{yan2019convolutional}
Yan, S., Li, Z., Xiong, Y., Yan, H., Lin, D.: Convolutional sequence generation
  for skeleton-based action synthesis. In: Proceedings of the IEEE/CVF
  International Conference on Computer Vision (ICCV) (2019)

\bibitem{yariv2020multiview}
Yariv, L., Kasten, Y., Moran, D., Galun, M., Atzmon, M., Ronen, B., Lipman, Y.:
  Multiview neural surface reconstruction by disentangling geometry and
  appearance. Advances in Neural Information Processing Systems (NeurIPS)
  \textbf{33} (2020)

\bibitem{zhou2019continuity}
Zhou, Y., Barnes, C., Lu, J., Yang, J., Li, H.: On the continuity of rotation
  representations in neural networks. In: Proceedings of the IEEE/CVF
  Conference on Computer Vision and Pattern Recognition (CVPR). pp. 5745--5753
  (2019)

\bibitem{zou20203d}
Zou, S., Zuo, X., Qian, Y., Wang, S., Xu, C., Gong, M., Cheng, L.: {3D} human
  shape reconstruction from a polarization image. In: Proceedings of the
  European Conference on Computer Vision (ECCV) (2020)

\end{thebibliography}
\newpage
\appendix
\section{Appendix}
\subsection{Training details}
\label{A:training}
The optimization is done with an Adam optimizer with a learning rate of 1e-3 for the MLP-based models and 1e-4 for the Transformer-based models.
We put some effort in tuning the learning rate for both models and settle on such simple settings to preserve comparability.
We train for 10000 epochs, but find that strong performance is often already reached earlier.
Unless otherwise stated we report performance for the model after 10000 epochs.
During training, we find that frequent sampling of representations is important for motion generation performance.
Thus for each training iteration,  we sample a sequence representation and an action representation per-sequence (as opposed to sampling a single action representation per-batch).
We set the weight for the Kullback-Leibler divergence to 1e-5, following the findings in \cite{petrovich21actor}.

Since the sequence and action representations are persistent parameters, we need to explicitly initialize them.
We set a unit log-variance for all MLP-based experiments and a log-variance to -10 for all Transformer-based experiments according to our findings in \cref{appendix:initialization}.

The full representation can be composed either through addition or concatenation of action-wise and sequence-wise representations.
For \textit{additive composition} we set the representation size of both action representation to $\alpha\in\mathbb{R}^{256}$ and sequence-wise representations to $\beta\in\mathbb{R}^{256}$ and the resulting sequence representation is given by $\alpha + \beta$.
For \textit{concatenation} we set both action representation to $\alpha\in\mathbb{R}^{128}$ and sequence-wise representations to $\beta\in\mathbb{R}^{128}$.
We investigate the effect of both approaches in \cref{appendix:composition} and find that the MLP-based decoder is best trained with composition through \textit{concatenation} and the Transformer-based decoder is best trained with \textit{additive composition}.

The optimization problems in Eqs. \ref{eq:model_train_objective} - \ref{eq:sequence_train_objective} require joint optimization with sequence-wise, action-wise and dataset-wise parameters.
We employ an alternating updating scheme where first all sequence-wise and action-wise parameters are updated with respect to a fixed model and then the model parameters are updated with respect to all other parameters.

Because our work produces persistent sequence representations, our decoder is trained with temporal embeddings $\tau$ corresponding to an absolute position within a groundtruth sample.
In other words, during training $\tau_0$ always corresponds to the element at $t=0$ of a real sample.
This is different from the previous work \cite{petrovich21actor}, which samples fixed-length subsequences during each training iteration.
In their case, $\tau_0$ corresponds to the first element of the subsequence that was input to the encoder.
As a consequence of training with random sub-sequences, the sequence representations of \cite{petrovich21actor} are entangled with the starting point of the sequence and randomly sampling a new sequence may produce a representation at any starting point including the middle of an ongoing action.
In our work, on the other hand, $\tau_0$ always corresponds to the beginning of an action.

Finally, the loss proposed by \cite{petrovich21actor} is memory-intensive due to the computation of a human mesh, which makes training on UESTC time-intensive.
Our experiments suggest that replacing the vertices loss with a joint loss akin to \cite{chuan2020action2motion} leads to similar performance and we utilize this loss for our experiments on UESTC.

In the kinematic tree of the skeleton representation, only the root joint can't be represented as a rotation.
We find that this can make it difficult to balance the root loss with the other joints, since the magnitude of the root joint is unconstrained.
For some actions with significant root motion (running, jumping) this may affect performance.
This may be mitigated by carefully weighing the root reconstruction loss against other losses or even using a distinct model just for the root joint.
However, for the sake of comparability to the previous work, we don't address this issue in our proposed method further.

\subsection{Generative Modeling}
\label{appendix:generativeModeling}
\begin{algorithm}[t]
\SetAlgoLined
\KwIn{set of sequence representations $C = \{c^i|i \in \mathcal{M}^z\}$; corresponding sequence lengths $T=\{T^i|i \in \mathcal{M}^z\}$; minimum interval size $d_{\min}$; minimum population $p_{\min}$; overlap $d_\text{overlap}$}
\KwOut{set of length intervals $l_t$}
$t_{\text{left}} \leftarrow \min(T)$\\
\While{$t_{\text{right}}<\max(T)$}{
    $t_{\text{right}} \leftarrow t_{\text{left}} + d_{\min} - d_\text{overlap}$\\
    $p = |\{T^i\in T | t_\text{left} \leq T^i \leq t_\text{right}\}|$\\
    \While{$p < p_{\min}$}{
        $t_\text{right} \leftarrow t_\text{right} + 1$\\
        $p = |\{T^i\in T | t_\text{left} \leq T^i \leq t_\text{right}\}|$\\
    }
    $t_{\text{left}} \leftarrow t_{\text{right}}$\\
    \If{$p\geq p_{\min}$}{
        $l_t \leftarrow l_t \cup \{(t_\text{left}, t_\text{right})\}$
    }
}
\tcp{last interval may not have the minimum population}
\While{$p < p_{\min} \text{ \& } t_\text{left}>0$}{
    $t_\text{right} \leftarrow t_\text{right} + 1$\\
    $p = |\{T^i\in T | t_\text{left} \leq T^i \leq t_\text{right}\}|$\\
}
$l_t \leftarrow l_t \cup \{(t_\text{left}, t_\text{right})\}$
\caption{Greedy interval fitting}
\label{appendix:alg_greedy_fitting}
\end{algorithm}
In the following we describe the details of \Cref{appendix:alg_greedy_fitting} as described in \cref{subsec:generative}. 
Identifying a suitable set of length intervals that cover all sequence lengths in the set of sequences of an action class is a combinatorial optimization problem akin to the Knapsack problem.
The objective is to maximize the number of intervals to allow fine-grained modeling of the sequence-length while also guaranteeing a minimum number of training samples within each interval and a minimum amount of overlap between intervals.

The algorithm is action-conditional and so it is only applied to sequence representations within an action class. The set of all representations with the same action label is denoted $\mathcal{M}^z$. The population of a set (cardinality) is denoted as $|\cdot|$.

The ability of the GMM to fit to complex data is determined by the number of components (mixtures) used and a larger number of components increases the chance of Gaussian component collapse, in which one of the components is identical to the distribution of a single sample.
If this happens, the sampling procedure just recreates training samples.
This type of collapse would inflate our performance under the realism metric, but is not desirable.
To avoid this, we first detect cases of collapsing by computing the distance between the mean of each Gaussian component and the representations of all training samples.
If we find a distance below a threshold (10\% of the average magnitude of all representation), we consider the GMM contain collapsed components.
Since the EM algorithm is sensitive to the initial conditions we repeat the fitting with different initial conditions until we find a non-collapsed distribution.
Typically we start with 15 components and if we can't find such a distribution after 100 attempts, we reduce the number of components and repeat the fitting procedure.

\subsection{Tools}
Our model is implemented with pytorch \cite{paszke2019pytorch}. For conversion between different rotation representations we use pytorch3D \cite{ravi2020pytorch3d} and for fitting GMMs to our representations we use scikit-learn \cite{scikit-learn}. Also we use the python implementation of smplify-x \cite{SMPL-X:2019} during loss computation and for visualization.

\subsection{Runtime}
We measure the inference time for the generation of a single 60 time-step sequence with our small MLP-based model ($7.53^{\pm0.22}$ ms), large MLP-based model ($7.66^{\pm0.21}$ ms) used on UESTC, and the Transformer-based model ($16.65^{\pm0.18}$ ms) and find that the MLP-based models are generally faster than Transformer-based models even with similar parameters counts.

Furthermore, we measure the time of a single training iteration with a batch size of 32, a temporal mini-batch of 5 and the reconstruction loss proposed by the Transformer baseline \cite{petrovich21actor}. A single iteration takes $115.53^{\pm0.19}$ ms on our small MLP-based model, $116.98^{\pm0.19}$ ms on our large MLP-based model and $138.49^{\pm0.78}$ ms on our Transformer-based model.

These time were measured on a single Nvidia Titan Xp under the same conditions. During actual training the batch size and temporal mini-batch size may differ and in particular training with larger datasets involves more sequence-wise parameters. We find that the training with our proposed settings of our MLP-based models takes 24 hours on HumanAct, 29 hours on NTU13 with a single Nvidia A100 and 40 hours on UESTC with a node of 5 Nvidia A100. As for our Transformer-based models, it takes 15 hours on HumanAct and 31 hours on NTU13 with a node of 4 Nvidia A100.

Note that during training our method needs to update each INR frequently.
This can be scaled to large datasets by distributing the INRs across multiple GPUs.
While we were able to train our light-weight MLP model on a large dataset such as UESTC, training a Transformer based model with the available resources would have been excessively slow. 
So, due to resource constraints, we didn’t perform full-scale experiments on UESTC with a Transformer-based model.
Given sufficient resources, training of a Transformer-based implicit model should be straightforward.

\subsection{Negative Societal Impacts}
\label{appendix:NegativeImpact}
The goal of this work is to generate life-like motions for animation or downstream tasks such as data augmentation.
Such life-like motions can contribute to the development of \textit{deep-fakes}, which could be used with malicious intend to impersonate or deceive.
With advances in monocular pose estimation, procuring motion data of people (and thus potential training data) without their knowledge or consent is becoming easier.
The authors strongly encourage research into systems that can detect augmented/fake and/or verify real media.

\subsection{Personal Data of Human Subjects}
\label{appendix:Privacy}
The HumanAct12, NTU RGBD and UESTC datasets are all publicly available, but no public statement about the consent of the human subjects is provided.
However, this work does not use personally identifiable information such as subject labels or appearance.

\subsection{Evaluation Metrics}
\label{A:EvaluationMetrics}
A difference between the evaluation of \cite{chuan2020action2motion} and \cite{petrovich21actor} is the frequency with which samples of a given action class are generated.
In \cite{chuan2020action2motion} motions of all action classes are generated equally often, irrespective of the frequency in the dataset.
This results in an inflated FID score, so we follow the protocol by \cite{petrovich21actor}, which generates motions with the frequencies found in the dataset.

The \textbf{Diversity} is computed by sampling two subset $\mathcal{S}_{d1}=\{f_1, ..., f_S\}$ and $\mathcal{S}_{d2}=\{\hat{f}_1, ..., \hat{f}_S\}$ of equal size $S_d$ from the features $f_i$ of all generated motions. We follow \cite{chuan2020action2motion} and use $\mathcal{S}_d=200$.
\begin{equation}
    \text{Diversity} = \frac{1}{S_d} \sum_{i=1}^{S_d} \left\|f_i-\hat{f_i}\right\|_2
\end{equation}
Similarly, the \textbf{Multimodality} is computed by sampling two subset $\mathcal{S}_{l1}=\{f_{z,1},$ $ ..., v_{f,S_l}\}$ and $\mathcal{S}_{l2}=\{\hat{f}_{z,1}, ..., \hat{f}_{z,S_l}\}$ of equal size $S_l$ from the features $v_{z,i}$ of the generated motions of action class $z$ averaged over all action classes. Again, we follow \cite{chuan2020action2motion} and use $\mathcal{S}_l=20$.
\begin{equation}
    \text{Multimodality} = \frac{1}{|\mathcal{M}_z|\cdot S_l}\sum_{z=1}^{\mathcal{M}_z}\sum_{i=1}^{S_l}\left\|f_{z,i} - \hat{f}_{z,i}\right\|
\end{equation}

\section{Additional Experiments}
\label{appendix:add_experiments}
\subsection{Initialization}
We find that the initialization of the variational implicit representations can effect the performance of our models. All representations are initialized with a zero mean and we scale the diagonal variance matrix with a scalar. In particular, our Transformer-based model fails to fit the training data accurately unless the variational representations are initialized with relatively small variance. On the other hand, our MLP-based model has reasonable performance independent of the initialization of the variance, but greater variances clearly improve performance.
\label{appendix:initialization}
\begin{table*}[h]
\begin{center}
\def\arraystretch{1.3}
\small
\begin{tabular}{c c c c c c}
     \hline
 \multicolumn{6}{c}{HumanAct12}\\
Method & Log-variance & $\text{FID}_{train}\downarrow$ & Accuracy $\uparrow$& Diversity $\rightarrow$ & Multimod. $\rightarrow$\\ 
 \hline\hline
 MLP & 1& $0.114^{\pm.001}$  &$0.970^{\pm.001}$ & $6.786^{\pm.057}$ & $2.507^{\pm.034}$\\
 MLP & 0& $0.163^{\pm.002}$  &$0.955^{\pm.001}$ & $6.868^{\pm.063}$ & $2.706^{\pm.034}$\\
 MLP &-10& $0.277^{\pm.004}$  &$0.881^{\pm.002}$ & $6.794^{\pm.059}$ & $3.340^{\pm.036}$\\
 \hline
 Transformer &1& $4.355^{\pm.022}$ & $0.536^{\pm.002}$ & $6.195^{\pm.053}$ & $3.619^{\pm.049}$\\
 Transformer &0& $1.812^{\pm.016}$ & $0.709^{\pm.003}$ & $6.559^{\pm.054}$ & $3.540^{\pm.037}$\\
 Transformer &-10& $0.088^{\pm.003}$ & $0.973^{\pm.001}$ & $6.881^{\pm.048}$ & $2.569^{\pm.040}$\\
 \hline
 \end{tabular}
 \caption{Ablation study for different initializations of the variational implicit neural representations. ($\pm$ indicates 95\% confidence interval, $\rightarrow$ closer to real is better}
 \label{appendix:tab:initialization}
\end{center}
\end{table*}
\subsection{Number of GMM components}
By using a conditional Gaussian Mixture Model (GMM) as a generative model, we can sample from a representation space that may be structured according to semantic factors such as action-class and sequence-length.
In the ablation study in \cref{tab:abl_gmm}, we investigate how the complexity of the GMM (determined by the number of components) affects the quality of our generated motions.
We find that our method is stable with respect to the number of components and even with a single component the method reaches the performance of the baseline Action2Motion\cite{chuan2020action2motion}.
By increasing the number of components of the GMM we can improve the realism, while maintaining diversity.
In particular, with 15 components the model can reach SOTA performance while maintaining a healthy Mean Maximum Similarity, which, as discussed in \cref{subsec:mean_maximum_similarity}, indicates that the model generates motions distinct from the training set.
\label{appendix:gmm_components}
\begin{table}
\small
\begin{center}
\def\arraystretch{1.3}
\begin{tabular}{c c c c c c}
     \hline
\multicolumn{6}{c}{HumanAct12}\\
\#& $\text{FID}\downarrow$ & Accuracy $\uparrow$& Diversity $\rightarrow$ & Multimod. $\rightarrow$ & MMS\\ 
 \hline\hline
1 & $0.351^{\pm.004}$  &$0.915^{\pm.002}$ & $6.761^{\pm.048}$ & $2.985^{\pm.040}$& $1.155^{\pm.005}$\\
3 &$0.209^{\pm.002}$&$0.943^{\pm.001}$&$6.755^{\pm.043}$&$2.753^{\pm.044}$& $1.015^{\pm.004}$\\
5 &$0.172^{\pm.003}$&$0.965^{\pm.001}$&$6.792^{\pm.041}$&$2.625^{\pm.031}$& $0.924^{\pm.005}$\\
10 &$0.125^{\pm.002}$&$0.971^{\pm.001}$&$6.792^{\pm.043}$&$2.698^{\pm.033}$& $0.902^{\pm.003}$\\
15 & $0.114^{\pm.001}$ & $0.970^{\pm.001}$ & $6.786^{\pm.057}$ & $2.507^{\pm.034}$ & $0.884^{\pm.004}$\\
 \hline
 \end{tabular}
 \caption{Ablation study of the number of components of each GMM for the MLP model on HumanAct.}
 \label{tab:abl_gmm}
 \end{center}
 \end{table}
 \subsection{Temporal Mini-Batch}
\label{appendix:temp_mini_batch}
During training we sample fixed-length, temporal mini-batches.
These can be sampled with two strategies; (R) randomly or as (C) consecutive subsequences.
Since our MLP-based model predicts each time-step independently, consecutive sampling isn't meaningful for it.
The Transformer-based model predicts multiple time-steps simultaneously, so here consecutive sampling is a reasonable choice.
We investigate the effect of different mini-batch sizes and sampling strategies.
Note, that all models were evaluated with a fixed evaluation length of 60 time-steps.
\begin{table*}[h]
\begin{center}
\def\arraystretch{1.3}
\small
\begin{tabular}{c c c c c c}
     \hline
\multicolumn{6}{c}{HumanAct12}\\
 Method &Batch size& $\text{FID}_{train}\downarrow$ & Accuracy $\uparrow$& Diversity $\rightarrow$ & Multimod. $\rightarrow$\\ 
 \hline\hline
 MLP & R5 & $0.114^{\pm.001}$  &$0.970^{\pm.001}$ & $6.786^{\pm.057}$ & $2.507^{\pm.034}$\\
 MLP & R10 &$0.141^{\pm.003}$&$0.965^{\pm.065}$&$6.856^{\pm.065}$&$2.634^{\pm.040}$\\
 Transformer & R5 & $0.214^{\pm.004}$ & $0.938^{\pm.001}$ & $6.755^{\pm.061}$& $2.877^{\pm.031}$\\
 Transformer & C5 & $0.498^{\pm.009}$ & $0.877^{\pm.002}$ & $6.674^{\pm.055}$&$3.208^{\pm.041}$\\
 Transformer & C60 & $0.088^{\pm.003}$ & $0.973^{\pm.001}$ & $6.881^{\pm.048}$ & $2.569^{\pm.040}$\\
 \hline
 \end{tabular}
 \caption{Ablation study for size of the temporal mini-batch. The temporal mini-batch is either constructed from random frames (R) or consecutive frames (C). ($\pm$ indicates 95\% confidence interval, $\rightarrow$ closer to real is better)}
 \label{appendix:tab:temp_mini_batch}
\end{center}
\end{table*}
\subsection{Representation Composition}
Our representations consist of sequence-wise representations and action-wise representations.
These are composed to become a single sequence representation by either addition or concatenation.
To ensure a fixed latent dimension of 256, we set the sequence-wise and action-wise representations to have 256 dimensions for additive composition and 128 dimensions when composing through concatenation.
In \cref{appendix:tab:composition} we investigate which composition performs best for which model and find that the MLP-based model performs best with composition through concatenation and the Transformer-based model performs best with additive composition.
\label{appendix:composition}
\begin{table*}[h]
\begin{center}
\def\arraystretch{1.3}
\small
\begin{tabular}{c c c c c c}
     \hline
  \multicolumn{6}{c}{HumanAct12}\\
 Method & Composition & $\text{FID}_{train}\downarrow$ & Accuracy $\uparrow$& Diversity $\rightarrow$ & Multimod. $\rightarrow$\\ 
 \hline\hline
 MLP & concat & $0.114^{\pm.001}$  &$0.970^{\pm.001}$ & $6.786^{\pm.057}$ & $2.507^{\pm.034}$\\
 MLP & add & $0.123^{\pm.003}$  &$0.960^{\pm.001}$ & $6.798^{\pm.058}$ & $2.642^{\pm.042}$\\
 \hline
 Transformer & concat & $0.090^{\pm.002}$ & $0.959^{\pm.001}$ & $6.861^{\pm.045}$ & $2.737^{\pm.026}$\\
 Transformer & add & $0.088^{\pm.003}$ & $0.973^{\pm.001}$ & $6.881^{\pm.048}$ & $2.569^{\pm.040}$\\
 \hline
 \end{tabular}
 \caption{Ablation study for different representation compositions. The sequence and action representations can be either added or concatenated. ($\pm$ indicates 95\% confidence interval, $\rightarrow$ closer to real is better)}
 \label{appendix:tab:composition}
\end{center}
\end{table*}
\section{Additional qualitative results}
We present additional qualitative results in \cref{appendix:fig:additional_results} as well as videos \footnote{Videos are included in the supplementary materials. Upon acceptance, they will be released.}. In \cref{appendix:fig:additional_results} we show motions generated by our MLP-based model with different target sequence-lengths. Generating short sequences in particular can be difficult, since the generated sequence should include the a full action, beginning to end. Our method is able to reliable generate such short complete actions due to our sequence-length conditional representation space. 

The videos are organized into skeleton videos for the HumanAct12 dataset and full 3D renderings for the UESTC dataset.
For the HumanAct12 dataset we provide a direct side-by-side comparison for Action2Motion\cite{chuan2020action2motion}, ACTOR\cite{petrovich21actor} and the proposed INR approach with a Transformer and MLP decoder.
For the UESTC dataset we provide a direct side-by-side comparison between ACTOR and the proposed INR approach with an MLP decoder.
To highlight the ability of each model to handle variable-length sequences, we generate sequences with sequences lengths of 40, 60 and 120 frames.
The sequences for the baselines were generated by the models provided by the authors without further fine-tuning.

\begin{figure*}[t]
    \begin{center}
    \includegraphics[width=0.8\textwidth, trim=0cm 0cm 0cm 0cm, clip]{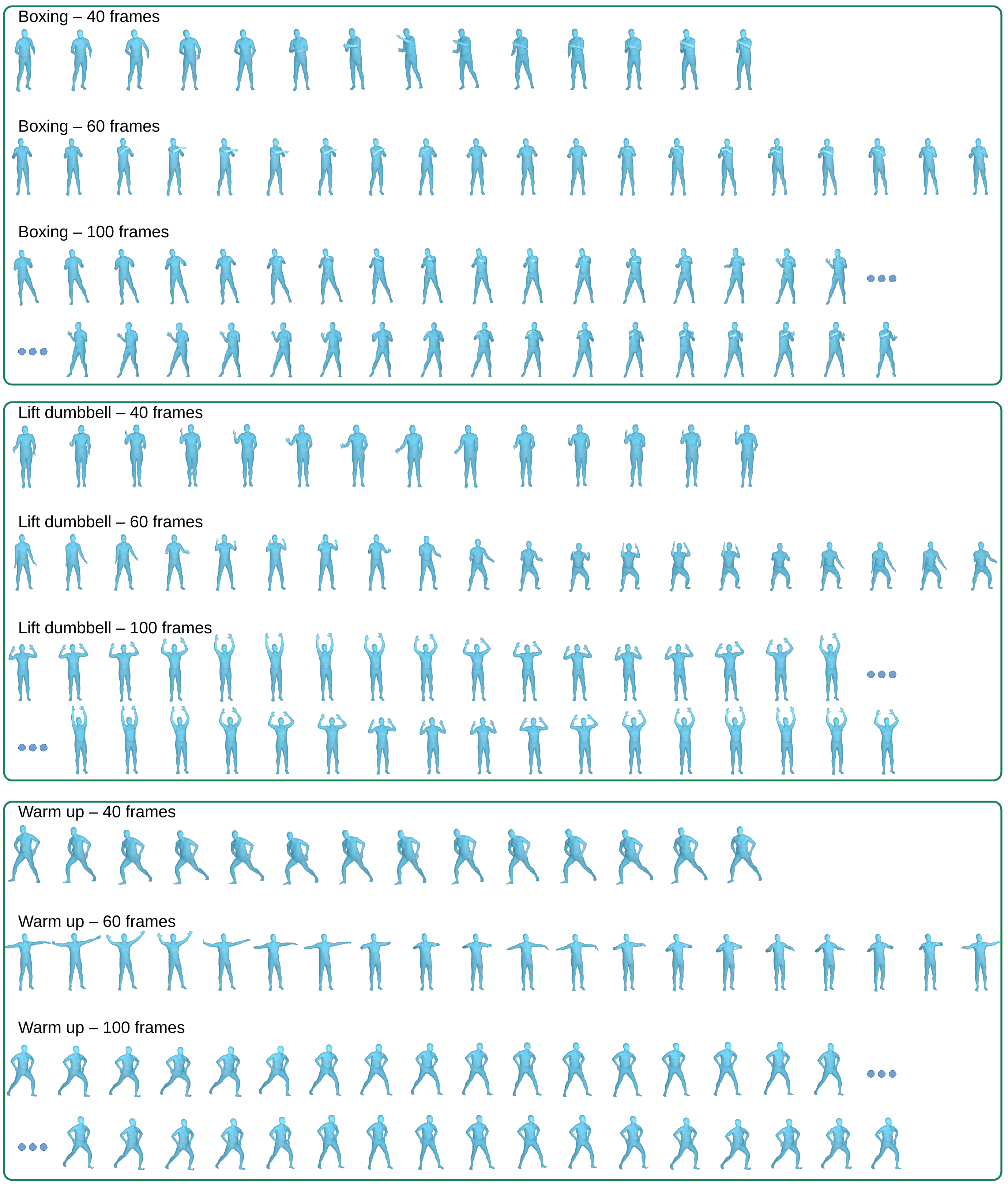}
    \caption{Examples of motions for the action classes \textit{boxing}, \textit{lift dumbbell} and \textit{warm up} generated with our MLP-based model with different target lengths. The generated motions are complete (finish within the target sequence length) and are diverse.}
    \label{appendix:fig:additional_results}
\end{center}
\end{figure*}
\label{A.AddQualitative}
\end{document}